\pdfoutput=1

\documentclass[11pt]{article}

\usepackage{EMNLP2022}
\usepackage{graphicx}

\usepackage{times}
\usepackage{latexsym}

\usepackage[T1]{fontenc}

\usepackage[utf8]{inputenc}

\usepackage{microtype}

\usepackage{inconsolata}

\title{DoSA : A System to Accelerate Annotations on Business Documents with Human-in-the-Loop}

\author{Neelesh K Shukla \and MSP Raja \and Raghu Katikeri \and Amit Vaid\\
        State Street Corporation \\ Bengaluru, KA, India \\ \texttt{\{nshukla, smaddila1, rkatikeri, avaid\}@statestreet.com}}

\begin{document}
\maketitle
\begin{abstract}
Business documents come in a variety of structures, formats and information needs which makes information extraction a challenging task. Due to these variations, having a document generic model which can work well across all types of documents and for all the use cases seems far-fetched. For document-specific models, we would need customized document-specific labels. We introduce DoSA (\textbf{Do}cument \textbf{S}pecific \textbf{A}utomated Annotations), which helps annotators in generating initial annotations automatically using our novel bootstrap approach by leveraging document generic datasets and models. These initial annotations can further be reviewed by a human for correctness. An initial document-specific model can be trained and its inference can be used as feedback for generating more automated annotations. These automated annotations can be reviewed by human-in-the-loop for the correctness and a new improved model can be trained using the current model as pre-trained model before going for the next iteration. In this paper, our scope is limited to Form like documents due to limited availability of generic annotated datasets, but this idea can be extended to a variety of other documents as more datasets are built. An open-source ready-to-use implementation is made available on GitHub. \footnote{https://github.com/neeleshkshukla/DoSA}
\end{abstract}

\section{Introduction}

With the recent advancements in technology and increased adoption of digitization, almost all organizations maintain and exchange business documents in digitized formats like PDFs, scans, faxes, images etc. These documents come in all shape, sizes and format like invoices, emails, medical reports, contracts, scientific papers and many more. The majority of the research has concentrated on documents present on the web that do not adequately capture the complexity of analysis or comprehension required for business documents. These documents require a multidisciplinary approach that includes understanding of layout and structure, computer vision, natural language processing. Usually, organizations rely on humans to manually process these documents. The ability to read, understand and interpret these documents is referred as Document Intelligence (DI) or Document Understanding. There have been recent advancements in this area specifically with deep learning where many architectures \cite{huang2022layoutlmv3, docFormer, donut} and annotated datasets \cite{docvqa, zhong2019publaynet, sroie, park2019cord} have been published for various DI tasks like Document Classification, Document Visual Q\&A, Form Understanding etc. These publicly annotated datasets mostly capture only few document types like receipts, scientific articles etc. It becomes challenging \& tedious to have annotated public datasets available for various document types which therefore brings in the very need of having customized annotated datasets for specific use cases and document types.

\begin{figure}
\centering
\setlength{\fboxrule}{0.5pt}
\setlength{\fboxsep}{0pt}
    \fbox{\includegraphics[width=55mm,scale=0.1]{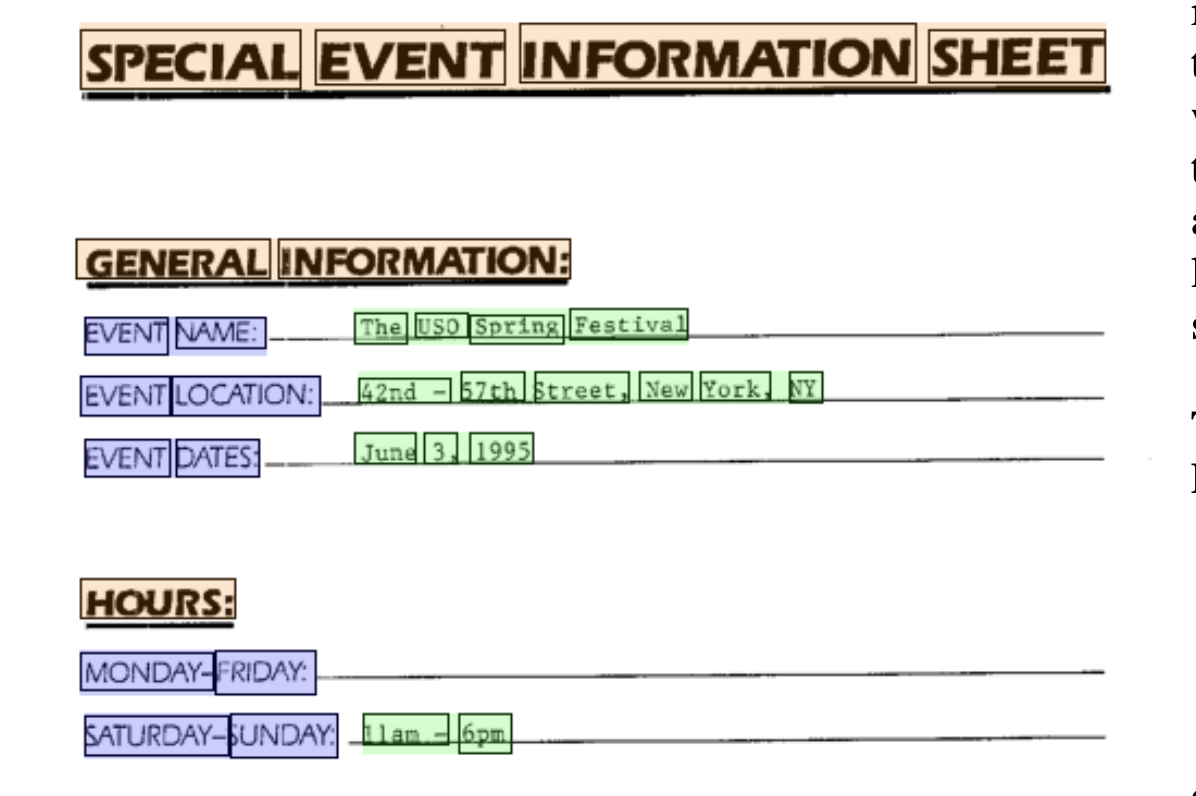}}
    \caption{Form Example from FUNSD: Keys are represented in blue, headers in orange, and Values in green.}
    \label{fig:funsd_sample}
\end{figure}

To limit our scope, in this paper we are focusing on information extraction from documents that follow a form-like structure. Forms are documents that have information usually present in Question-Answer or Key-Value format as shown in Figure \ref{fig:funsd_sample}. Documents like invoices, driving licenses, passports, medical records, financial statements, tax forms, quotations, payment cards, etc. fall under this category. 

There has been an effort in building a generic form dataset FUNSD\footnote{https://guillaumejaume.github.io/FUNSD/} \cite{jaume2019funsd}. FUNSD is a dataset for form understanding in noisy scanned documents that aim at extracting
and structuring the textual content of forms. It proposes an idea where a generic document can be represented via generic information and labels like question (or key), answer (or value), header and others. A model can consume this kind of generic representation to extract generic information. In most of the scenarios, users are interested in document-specific meaningful labels like document number, document date, etc and extracting a subset of information. Having a document generic labeling approach results in a noisy and verbose extraction. Therefore a need for document-specific annotations arises.
Commercial solutions like Microsoft Form Recognizer\footnote{https://azure.microsoft.com/en-in/services/form-recognizer/} and Google Document AI \footnote{https://cloud.google.com/document-ai} mostly support specific document type pre-built models and provide a facility to custom train a model for specific documents\footnote{https://docs.microsoft.com/en-us/azure/applied-ai-services/form-recognizer/concept-custom} which require new annotations. The SOTA models of document intelligence use multimodality of the document: text, position and image. Due to a change in the format or layout of the document, these modalities might be affected and a new model needs to be trained which will need a new set of annotations. With these, manual annotation becomes repetitive, laborious, expensive and time-consuming.

To reduce the time and human effort, we are proposing an active learning based automated annotation system DoSA (\textbf{Do}cument \textbf{S}pecific \textbf{A}utomated Annotations), where the initial set of document-specific annotations are generated by the system which can be reviewed by human annotators for correctness. An initial model can be trained with these annotations and its inference can be taken by the system as feedback to generate annotations on new documents and improve the model incrementally with the human in the loop.

\begin{figure*}
\centering
\setlength{\fboxrule}{0.5pt}
\setlength{\fboxsep}{0pt}
    \fbox{\includegraphics[width=\textwidth,scale=0.1]{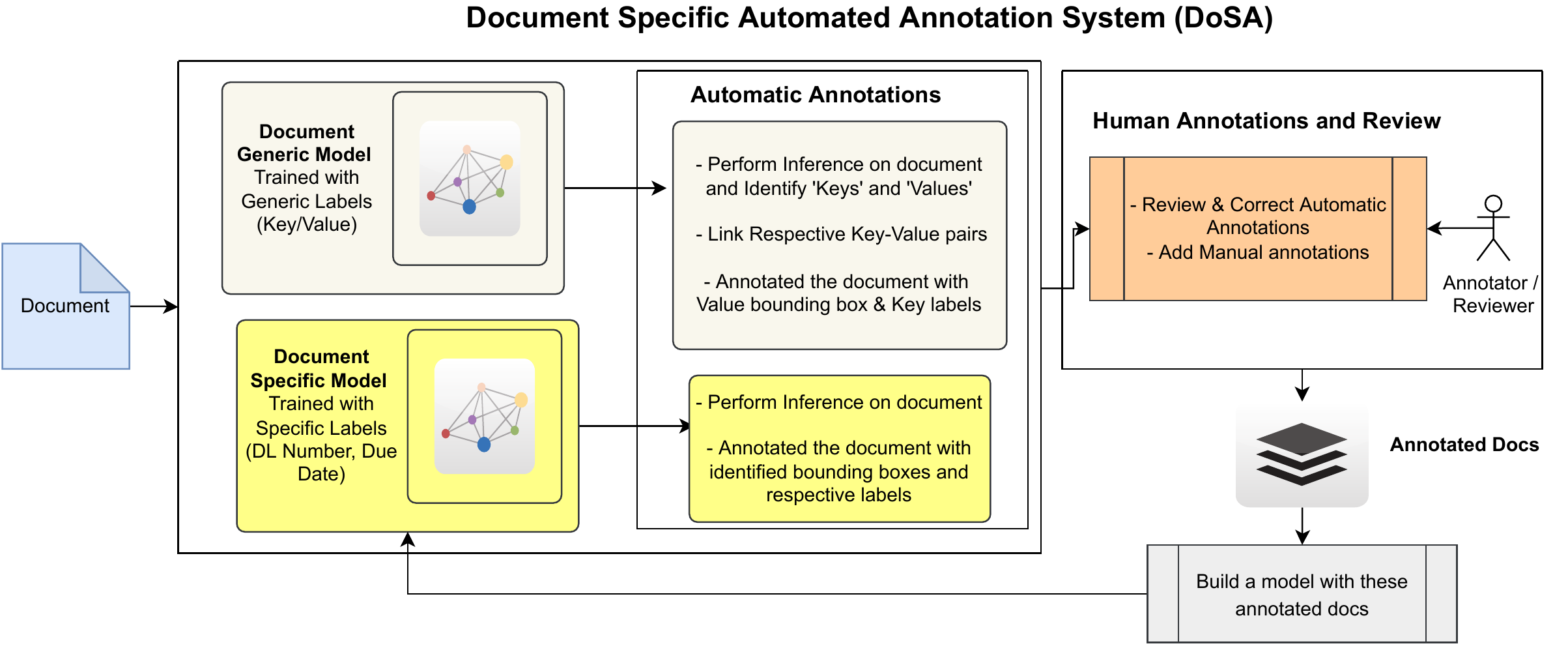}}
    \caption{DoSA System Overview: Generating Annotations with Human in the Loop}
    \label{fig:dosa_overview}
\end{figure*}

The main contribution of this paper is a novel bootstrapping approach to generate automated document-specific labels. To the best of our knowledge, this is the first attempt at generating automated annotations on business documents that contains visual and layout structure information along with the text. All other previous approaches have mostly focused on web or text \cite{10.1145/775152.775178, 10.1145/3230665} or images \cite{8991727}. We have seen an attempt on research documents but the scope was limited to automatically annotating documents with topic \cite{6690015}.

\section{DoSA System}

A high-level flow of the DoSA system is shown in Figure \ref{fig:dosa_overview}, where initial document-specific annotations are generated by document-generic model which is later reviewed by a human for correctness. With these initial annotations, a document-specific model is trained and its inference is taken as feedback to annotate further documents. A human will again review and correct the annotations and the reviewed documents can be added back to training for further improving the model. As the model matures, eventually the user would end up correcting a minimal number of annotated fields/documents. The journey for the model to get more precise with less human feedback is achieved by employing active learning strategies like uncertainty-based sampling which improve the document-specific model performance in very few iterations.

\subsection{Bootstrapping: Generating Initial Document Specific Annotations with Document Generic Model}
Here are the steps that are followed in generating annotations on the initial set of documents:
\begin{itemize}
    \item A document is processed via an OCR engine to get the words and their respective bounding boxes. DoSA uses open source OCR engine pytesseract\footnote{https://pypi.org/project/pytesseract/}.
    \item As an intermediate step, the text areas are identified as 'Keys' and 'Values' in these documents  (Section \ref{section:dosa_key_value_annotations}).
    \item Link respective Key and Values and form a key-value pair <K, V> (Section \ref{section:dosa_key_value_linking}).
    \item Document-specific annotations can be generated by labeling the area identified as value V with the text of the area identified by the respective key K (Section \ref{section:dosa_document_specific_annotations}). 
    
\end{itemize}
\subsubsection{Locating Keys and Values}
\label{section:dosa_key_value_annotations}
As an intermediate step, text areas in a document are identified as 'Key' and 'Value'. DoSA uses LayoutLMv3\cite{huang2022layoutlmv3} model for entity/token classification fine-tuned on generic FUNSD dataset\footnote{https://huggingface.co/nielsr/layoutlmv3-finetuned-funsd}. This can classify the word/token/entity in 'Key (Question)', 'Value (Answer)', 'Header' and 'Others' with F1 0.9078.
The areas are represented by bounding box coordinates <x1,y1, x2, y2> which are used for comparing the position and drawing rectangles in the next sections.  In the fax cover example shown in figure \ref{fig:dosa_example_key_value_annotations}, 'To', 'Fax Number', 'Phone Number', 'Date' has been identified as keys and 'George Baroody', '(336) 335-7392', '12/10/98' as values.

\subsubsection{Key and Value Pair Linking}
\label{section:dosa_key_value_linking}
After identifying keys and values, DoSA links respective pairs <K, V>. There have been multiple recent works addressing this Form Entity Linking problem \cite{structext, zhang-etal-2021-entity}. These works have F1 0.4 and 0.64 respectively. These SOTA models are not good enough and were resulting in a lot of noisy pairs. Based on manual observations of a few documents, we designed the following heuristics to identify key-value pairs.

Given a list of values V ordered by their position in a document, Value Vj is linked to candidate key Ki if it satisfies the following conditions:\\
\textbf{H1:} Position of Ki is less than the position of Vj.\\
\textbf{H2:} Ki is not linked to any other Value Vk\\
\textbf{H3:} Ki is the closest to Vj compared to other candidate keys Km which satisfy H1 and H2.\\

\begin{figure}
\centering
\setlength{\fboxrule}{0.1pt}
\setlength{\fboxsep}{1pt}

    \fbox{\includegraphics[width=75mm,scale=0.2]{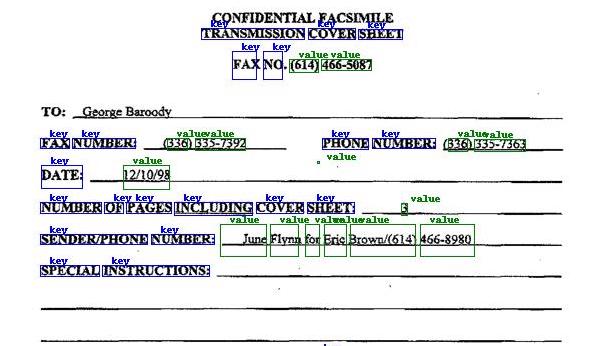}}

    \caption{Areas marked with generic 'Key' and 'Value' labels as described in section \ref{section:dosa_key_value_annotations}}
    \label{fig:dosa_example_key_value_annotations}
\end{figure}

If No such key is found for a Value Vj. Vj will be dropped else pair <Ki, Vj> is added to the output. For the document shown in figure \ref{fig:dosa_example_key_value_annotations}, some of the examples are <Fax Number: (336) 335-7392> and <Date: 12/10/98>.

\begin{figure}
\centering
\setlength{\fboxrule}{0.1pt}
\setlength{\fboxsep}{1pt}
    \fbox{\includegraphics[width=75mm,scale=0.2]{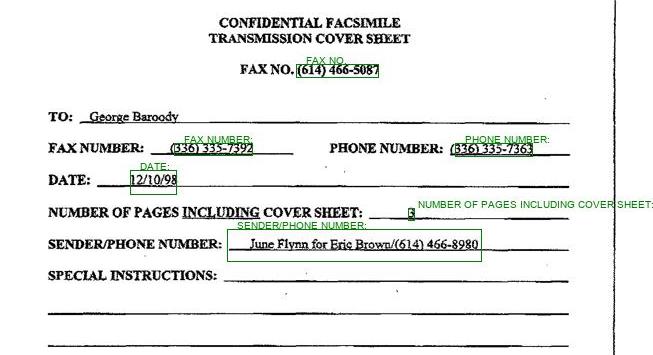}}
    \caption{Document specific annotations by labeling regions/texts identified as 'Value' with respective 'Key' in intermediate annotations as shown in figure \ref{fig:dosa_example_key_value_annotations}.}
    \label{fig:dosa_example_final_annotations}
\end{figure}

\subsubsection{Document Specific Annotations and Review with Human-in-the-Loop}
\label{section:dosa_document_specific_annotations}
Once the <Key, value> pairs are identified, annotations can be generated by drawing the bounding boxes around 'value' and annotating it with the text of the respective 'key'. These automated annotations can be submitted for review and modifications with Human-in-the-loop.

An example is shown in Figure \ref{fig:dosa_example_final_annotations}, once the key-value pair <Fax Number: (336) 335-7392> identified, the value (336) 335-7392 has been annotated with respective key 'Fax Number'.

\subsection{Annotations with Document Specific Model}
A custom initial model can be built by fine-tuning the generic model used in section \ref{section:dosa_key_value_annotations} on these reviewed initial annotations which now have document-specific labels. This document-specific model can be used to generate annotations via inference feedback on new documents. As more documents and annotations are added and reviewed, the model will eventually get mature.

\section{Conclusion and Future Work}
In this work, we presented DoSA, a system to generate document specific annotations from model built on document generic datasets. Our scope was limited to Form like document which can be further enhanced with the availability of new type of generic datasets. This system in current state can only take one type of document to generate one set of annotations. In case the users have multiple type of documents, they have to group the documents by type beforehand and use this system for individual groups. A layer can be added on top of DoSA system to automatically classify the documents and use DoSA for individual groups. As this work is still in progress, in this paper we focused on proposing this idea. We are planning to discuss the effectiveness of our proposed approaches and overall system in the near future.

\bibliography{anthology,custom}
\bibliographystyle{acl_natbib}

\end{document}